# Automatic symmetry based cluster approach for anomalous brain identification in PET scan image – An Analysis

A.Meena[1,] K. Raja[2]

[1]Research Scholar, Dept. of CSE, Sathyabama University, Chennai
[2]Principal, Narasu's Sarathy Institute of Technology, Salem

[1]kabimeena2@hotmail.com
[2]raja_koth@yahoo.co.in

**Abstract**

Medical image segmentation is referred to the segmentation of known anatomic structures from different medical images. Normally, the medical data researches are more complicated and an exclusive structures. This computer aided diagnosis is used for assisting doctors in evaluating medical imagery or in recognizing abnormal findings in a medical image. To integrate the specialized knowledge for medical data processing is helpful to form a real useful healthcare decision making system.

This paper studies the different symmetry based distances applied in clustering algorithms and analyzes symmetry approach for Positron Emission Tomography (PET) scan image segmentation. Unlike CT and MRI, the PET scan identifies the structure of blood flow to and from organs. PET scan also helps in early diagnosis of cancer and heart, brain and gastro-intestinal ailments and to detect the progress of treatment. In this paper, the scope diagnostic task expands for PET image in various brain functions.

## 1. Introduction

The most commonly used radiographic techniques are known as Computed Tomography (CT), Magnetic Resonance Imaging (MRI) and Positron Emission Tomography (PET). These technologies are major component techniques in diagnosis, clinical studies, treatment planning and are widely used for medical research. The motive of automatic medical image segmentation is to describe the image content based on its features. In recent years an ample of approaches has been proposed to segment medical images according its merits and limitations. The symmetry based approach is mathematically defined as a distance preserving transformation of the plane or space which leaves a given set of points unchanged and their respective features.

A Positron Emission Tomography (PET) is also known as PET imaging or a PET scan, is a type of nuclear medicine imaging. This scan image detects chemical and physiological changes related to metabolism. It uses a radioactive substance (tracer) to look for disease in the body. This scan images are more sensitive than other image techniques such CT and MRI.

This paper presents a study on the application of clustering to exploit the symmetry inherently present in the objects. It presents a collection of various research works in this area and highlights the contribution of these research papers focus on medical image segmentation exclusively brain functions.

## 2. Clustering

Clustering is used to classify items into similar groups in the process of data mining. It also exploits segmentation which is used for quick bird view for any kind of problem. Unlike classification, clustering and unsupervised learning do not rely on predefined classes and class labeled training examples. For this reason, clustering is a form of learning by observation, rather than learning by examples.

Many clustering techniques have been proposed over the years from different research disciplines. These techniques are used to perform with a given data and are being applied in an ample variety of interdisciplinary applications. For example, clustering can be used to derive plant and animal taxonomies in the field of biology; categorize genes with similar functionality in genetic engineering and clustering can be used in business to group the population by demographic information into segments for direct marketing and sales.

### 3. Metrics and Measures of Clustering

Cluster measure is used to calculate the similarity between pairs of clustering. The basic step in cluster formation is to calculate the distance between two datasets, which will identify the similarity of two elements. This will influence the shape of the clusters, as some elements may be close to one another according to one distance and farther away according to another. Euclidean distance is probably the most common chosen type to calculate distance between two datasets based on similarity. Smaller ED means better similarity and vice versa.

Validation is used to identify the optimal number of clusters and to measure the quality from other algorithms or using different parameters in same. The result of clustering must be validated in every application.

### 4. Symmetry

Symmetry is a pre-attentive feature which improves recognition and reconstruction of shapes and objects. Point symmetry and line symmetry are two keynotes of the geometrical symmetry [1]. In nature symmetry is a basic feature of shapes and objects. Instead of using similarity measure symmetry is also used to find the clusters in the feature space.

*4A. Symmetry Types*

The property of an object that enables it to undergo certain manipulations is called as symmetry operations.

An image that has point symmetry is unchanged in appearance by a 180 degree rotation. To determine whether an image has point symmetry is to turn it upside-down and see if it looks the same. The reflection symmetry is the imaginary line to fold the image and have both halves match exactly.

*4 B. Symmetry and distance metrics*

A new type of non metric distance based on point symmetry distance in [2], [3]. Pattern classification is done for a particular cluster depending on the symmetry based distance $ds$ rather than ED, only when $ds$ is greater than some user specified threshold θ.

The clusters are symmetrical to each other with respect to a number of intermediate points. But in some real data sets, which will fail to detect the proper number of clusters. Most of the validity measure usually assumes a certain geometrical structure in the cluster shape. But if several different structure s exist in the same data set, these have often been found to fail. To overcome these limitations of the Ps-index, a new cluster validity index named S-index is proposed in [6].

Determining the appropriate number of clusters from a given data set is an important consideration in clustering. A newly proposed point symmetry based cluster validity index, Sym-index, is used as a measure of the validity of the corresponding partitioning. The algorithm is therefore able to detect both convex and non-convex clusters irrespective of their sizes and shapes as long as they possess the symmetry property [7].

*4 C. Genetic algorithm based clustering and Symmetry*

Genetic algorithm based clustering is used to avoid the local optimum solution by optimizing the symmetry based distance rather than Euclidean Distance [5].

A line-symmetry-based distance is proposed in [8]. Adaptive mutation and crossover operations are used in genetic algorithm line symmetry distance. In real life application this algorithm is used for locating human faces from an image.

### 5. MR Medical Image Segmentation

An analysis of medical images for computer aided diagnosis and treatment pixel classification as a preliminary processing task. Medical image segmentation is a complex and challenging task due to the basically inexact nature of the images. Successful numerical results in segmented medical image can help physicians and neurosurgeons to investigate and diagnose the structure and function of the body in both health and disease [12]. Fully automatic brain tissue classification from magnetic resonance images (MRI) is of great importance for research and clinical study of much neurological pathology.

Clustering approaches have been widely used for segmentation of MR brain image [9]. In this paper, the first principle axis is determined by PCA method. Then, the newly developed line-sym index is used in VGALS approach. This method is helpful to solve intensity space problem and to measure the fitness of chromosome. But in certain cases, the obtained result is incorporated with some spatial information. Fig.1 has shown the segmentation of pixels in brain using MR scan image.

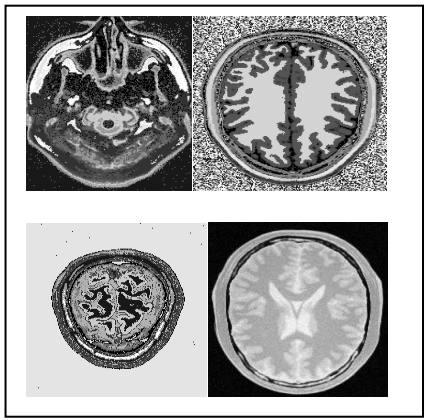

Fig.1 Pixel segmentation in MRI brain images using VGALS approach

## 6. PET Scan Image Segmentation

A PET image will display in different levels of positrons according to brightness and color. The images are reconstructed or interpreted by computer analysis and it is used for diagnosing cancer and assessing its spread.

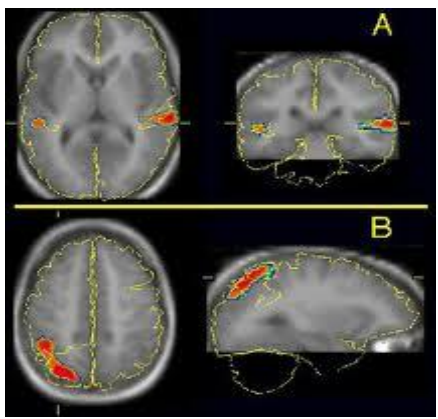

Figure 2: Brain image derived from PET scan

This image also can be used to distinguish between benign and malignant disease, establishing the extent of spread, measuring response to therapy and locating primary and secondary sites. Fig.2 shows the brain image from PET scan

The PET scan image has a number of benefits over other form of medical image such as X –rays, computed tomography (CT) and magnetic resonance imaging (MRI). These are useful tools using Anatomical or non - invasive imaging procedure that measures change in body structure. These procedures are limited for a number of diseases, particularly in disease that affect the brain. PET is a unique Metabolic or invasive imaging procedures based on molecular biology. PET scan reveals the cellular level metabolic changes occurring in an organ or tissue before anatomical or structural changes occur. It is an important and unique one often begins with functional changes at the cellular level. A PET scan can often detect these very early changes whereas a CT or MRI detect changes a little later as the disease begins to cause changes in the structure of organs or tissues. The major role of PET technology is used to treat neurological disorders such as Alzheimer's disease (AD), and other dementias. It is a serious loss of thinking ability in a person. In 2006, there were 26.6million sufferers worldwide. Alzheimer's is predicted to affect 1 in 85 people globally by 2050[15].

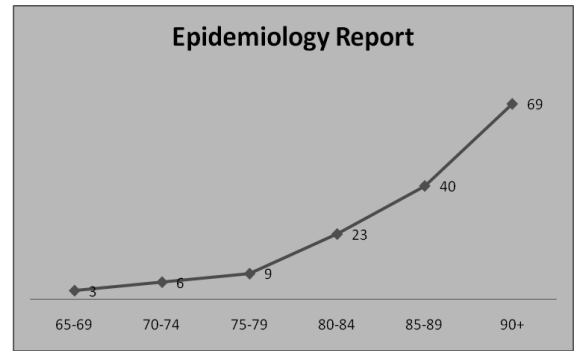

Fig.3 Affected person rate after age 65

Fig.3 has shown number of new cases affected per thousand person–years. Advancing age is a primary risk factor for the disease and incidence rates are not equal for all ages. Every five years after the age of 65, the risk of acquiring the disease approximately doubles, increasing from 3 to as much as 69 per thousand person years. Normally women are having a higher risk of developing AD [16].

## 7. Clinical Studies of Brain functions Using PET

The mixture of oxygen and glucose is normally known as the source of energy for brain. Additional chemical component phospholipids cover the nerve system and transfer the information to other cell. Some brain diseases like Alzheimer's disease and brain tumors affect the flow of blood to brain. The most common image tool such as MRI and CT scan doesn't cause major changes to diagnose this disease. But PET imaging helps the physicians and researchers to understand how the brain works in health and illness. It is able to identify in radiant color the regions of brain where nerve cells are working during a certain mental task.

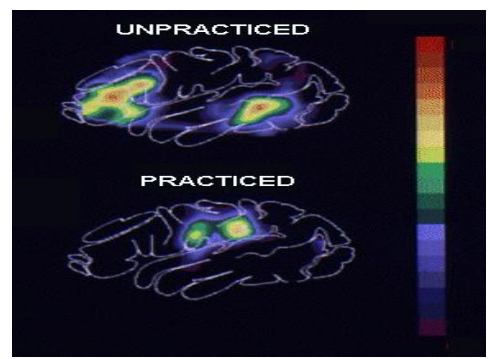

Fig.4 Experimental and clinical studies of brain function using PET

The fig.4 shown in the images made by Dr. Marcus Raichle, at the Neuroimaging Lab or the Washington University School of Medicine, St Louis, USA. An individual hearing a new language is shown in uppermost and the lowermost shows the same individual has now learned the language task and is spelling out. The various color tones demonstrates the working principles of cell activities with a higher energy and supplementary blood flow. Yellow and red regions are "hotter" and blue and black regions show decreased activity or none at all. The hotter color tone information is helpful to make a better diagnosis or surgical intervention, if needed.

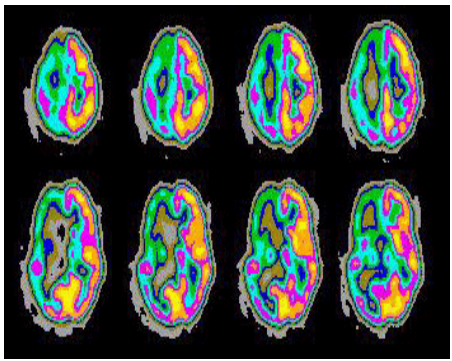

Fig.5 PET image of Epileptic seizure caused by pathological conditions

Epilepsy is a kind of disorder which can be represented as seizures. The causes of epilepsy can be determined mainly on the basis of communication problem between the nerve cells in brain. It is considered to be mutations in several genes that have been linked to some types of epilepsy and children born with an abnormal structure of brain. The fig.5 has shown the various stages of epilepsy occurred by some pathological conditions.

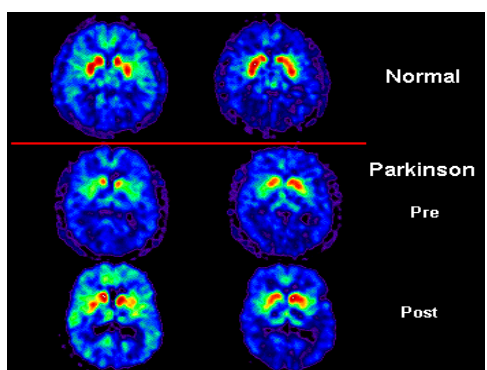

Fig. 6 Neurological alterations from PET scan device

Basal ganglia is a part of brain which formed DOPA a chemical substance. It is necessary for the functional activities of brain. If the level of DOPA is reduced may cause the neurological alterations called Parkinson's disease. The fig.6 shows the conditional stages of Parkinson's made in PET of the Crump Institute for Biological Imaging, University of California at Los Angeles.

## 8. Conclusion

Medical image analysis is prominently used by physicians and doctors to investigate and diagnose of structure and function of the body. This paper provides symmetry based automatic pixel segmentation approach for PET scan images. This method does not require any prior knowledge about the number of clusters existing in the image. Therefore, it reduces the computational complexity. Further it focuses on PET scan medical image processing which is used to treat neurological disorders known as AD. This PET scan image classification will be implemented in symmetry based clustering technique and to compare the results in other non-invasive imaging techniques.